\documentclass{article} 
\usepackage{iclr2019_conference,times}

\usepackage{amsmath,amsfonts,bm}

\def\eqref#1{equation~\ref{#1}}

\def\1{\bm{1}}

\DeclareMathAlphabet{\mathsfit}{\encodingdefault}{\sfdefault}{m}{sl}
\SetMathAlphabet{\mathsfit}{bold}{\encodingdefault}{\sfdefault}{bx}{n}

\usepackage{hyperref}
\usepackage{url}
\usepackage{algpseudocode,algorithm,algorithmicx}
\usepackage{graphicx}
\usepackage{xcolor}
\usepackage{amssymb}
\usepackage{tikz}
\usepackage{authblk} 
\usepackage{comment}

\newcommand{\tompson}[1]{\ifdefined\DRAFT \textcolor{orange}{(tompson: #1)} \else \fi}

\newcommand{\kostrikov}[1]{\ifdefined\DRAFT \textcolor{red}{(kostrikov: #1)} \else \fi}
\newcommand{\debidatta}[1]{\ifdefined\DRAFT \textcolor{brown}{(debidatta: #1)} \else \fi}

\title{Discriminator-Actor-Critic: \\
Addressing Sample Inefficiency and Reward Bias in Adversarial Imitation Learning}

\author[1,2,*]{Ilya Kostrikov}
\author[2, $\dagger$]{Kumar Krishna Agrawal}
\author[2, $\dagger$]{Debidatta Dwibedi}
\author[2]{Sergey Levine}
\author[2]{Jonathan Tompson}
\affil[1]{Courant Institute of Mathematical Sciences, New York University, New York, NY}
\affil[2]{Google Brain, Mountain View, CA}
\affil[*]{Work done as an intern at Google Brain}

\iclrfinalcopy 
\begin{document}

\maketitle

\begin{abstract}
We identify two issues with the family of algorithms based on the Adversarial Imitation Learning framework. The first problem is implicit bias present in the reward functions used in these algorithms. While these biases might work well for some environments, they can also lead to sub-optimal behavior in others. Secondly, even though these algorithms can learn from few expert demonstrations, they require a prohibitively large number of interactions with the environment in order to imitate the expert for many real-world applications. In order to address these issues, we propose a new algorithm called Discriminator-Actor-Critic that uses off-policy Reinforcement Learning to reduce policy-environment interaction sample complexity by an average factor of 10. Furthermore, since our reward function is designed to be unbiased, we can apply our algorithm to many problems without making any task-specific adjustments. 

\end{abstract}


\section{Introduction}

The Adversarial Imitation Learning (AIL) class of algorithms learns a policy that robustly imitates an expert's actions via a collection of expert demonstrations, an adversarial discriminator and a reinforcement learning method. For example, the Generative Adversarial Imitation Learning (GAIL) algorithm~\citep{ho2016generative} uses a discriminator reward and a policy gradient algorithm to imitate an expert RL policy on standard benchmark tasks. 
Similarly, the Adversarial Inverse Reinforcement Learning (AIRL) algorithm~\citep{fu2017learning} makes use of a modified GAIL discriminator to recover a reward function that can be used to perform Inverse Reinforcement Learning (IRL)~\citep{abbeel2004apprenticeship} and who`s subsequent dense reward is robust to changes in dynamics or environment properties. 
Importantly, AIL algorithms such as GAIL and AIRL, obtain higher performance than supervised Behavioral Cloning (BC) when using a small number of expert demonstrations; experimentally suggesting that AIL algorithms alleviate some of the distributional drift~\citep{ross2011reduction} issues associated with BC. 
However, both these AIL methods suffer from two important issues that will be addressed by this work: 1) a large number of policy interactions with the learning environment is required for policy convergence and 2) bias in the reward function formulation and improper handling of environment terminal states introduces implicit rewards priors that can either improve or degrade policy performance.

While GAIL requires as little as 200 expert frame transitions (from 4 expert trajectories) to learn a robust reward function on most MuJoCo~\citep{todorov2012mujoco} tasks, the number of policy frame transitions sampled from the environment can be as high as 25 million in order to reach convergence. If PPO~\citep{schulman2017proximal} is used in place of TRPO~\citep{schulman2015trust}, the sample complexity can be reduced somewhat (for example, as in Figure \ref{fig:results}, 25 million steps reduces to approximately 10 million steps), however it is still intractable for many robotics or real-world applications. In this work we address this issue by incorporating an off-policy RL algorithm (TD3~\citep{fujimoto2018addressing}) and an off-policy discriminator to dramatically decrease the sample complexity by many orders of magnitude.

In this work we will also illustrate how the specific form of AIL reward function used has a large impact on agent performance for episodic environments. For instance, as we will show, a strictly positive reward function prevents the agent from solving tasks in a minimal number of steps and a strictly negative reward function is not able to emulate a survival bonus. Therefore, one must have some knowledge of the true environment reward and incorporate such priors to choose a suitable reward function for successful application of GAIL and AIRL. We will discuss these issues in formal detail, and present a simple - yet effective - solution that drastically improves policy performance for episodic environments; we explicitly handle absorbing state transitions by learning the reward associated with these states.

We propose a new algorithm, which we call Discriminator-Actor-Critic (DAC), that is compatible with both the popular GAIL and AIRL frameworks, incorporates explicit terminal state handling, an off-policy discriminator and an off-policy actor-critic reinforcement learning algorithm. DAC achieves state-of-the-art AIL performance for a number of difficult imitation learning tasks. More specifically, in this work we:

\begin{itemize}
\item Identify, and propose solutions for the problem of bias in discriminator-based reward estimation in imitation learning.
\item Accelerate learning from demonstrations by providing an off-policy variant for AIL algorithms, which significantly reduces the number of agent-environment interactions.
\item Illustrate the robustness of DAC to noisy, multi-modal and constrained expert demonstrations, by performing experiments with human demonstrations on non-trivial robotic tasks.
\end{itemize}

\begin{figure}
\centering
\includegraphics[width=0.95\linewidth]{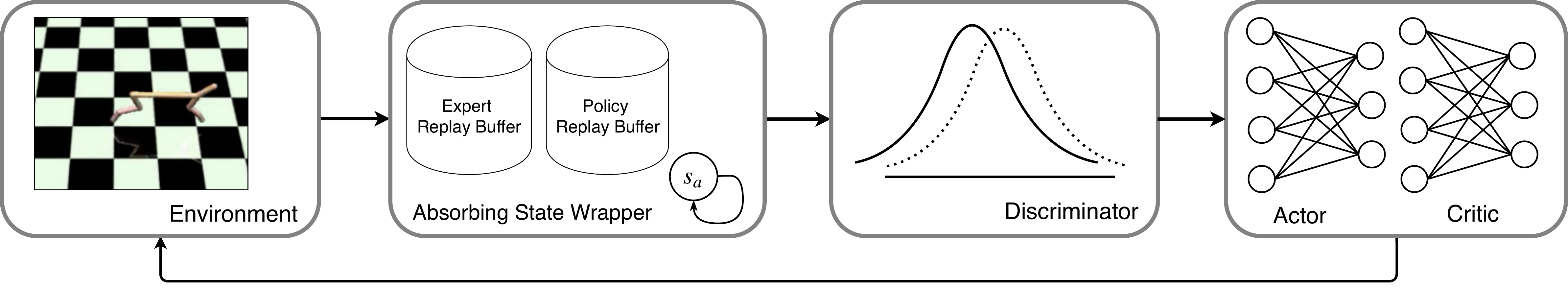}
\caption{The Discriminator-Actor-Critic imitation learning framework.}
\label{fig:opgail}
\end{figure}

\section{Related Work}


Imitation learning has been broadly studied under the twin umbrellas of Behavioral Cloning (BC) \citep{bain1999framework, ross2011reduction} and Inverse Reinforcement Learning (IRL) \citep{Ng00algorithmsfor}. To recover the underlying policy, IRL performs an intermediate step of estimating the reward function followed by RL on this function \citep{abbeel2004apprenticeship, ratliff2006maximum}. Operating in the Maximum Entropy IRL formulation~\citep{ziebart2008maximum}, \cite{finn2016guided} introduce an iterative-sampling based estimator for the partition function, deriving an algorithm for recovering non-linear reward functions in high-dimensional state and action spaces. \cite{finn2016connection} and \cite{fu2017learning} further extend this by exploring the theory and practical considerations of an adversarial IRL framework, and draw connections between IRL and cost learning in GANs \citep{goodfellow2014generative}.

In practical scenarios, we are often interested in recovering the expert's policy, rather than the reward function.
Following \cite{syed2008apprenticeship}, and by treating imitation learning as an occupancy matching problem, \cite{ho2016generative} proposed a Generative Adversarial Imitation Learning (GAIL) framework for learning a policy from demonstrations, which bypasses the need to recover the expert's reward function. More recent work extends the framework by improving on stability and robustness \citep{wang2017robust, kim2018imitation} and making connections to model-based imitation learning \citep{baram2017end}. These approaches generally use on-policy algorithms for policy optimization, trading off sample efficiency for training stability.





Learning complex behaviors from sparse reward signals poses a significant challenge in reinforcement learning. In this context, expert demonstrations or template trajectories have been successfully used \citep{peters2008reinforcement} for initalizing RL policies. There has been a growing interest in combining extrinsic sparse reward signals with imitation learning for guided exploration \citep{zhu2018reinforcement, kang18a, le2018hierarchical, vecerik2017leveraging}.
Off policy learning from demonstration has been previously studied under the umbrella of accelerating reinforcement learning by structured exploration \citep{nair2017overcoming, hester2017deep} An implicit assumption of these approaches is access to demonstrations and reward from the environment; our approach requires access only to expert demonstrations.



Our work is most related to AIL algorithms \citep{ho2016generative, fu2017learning, torabi2018generative}. In contrast to \cite{ho2016generative} which assumes \textit{(state-action-state')} transition tuples, \cite{torabi2018generative} has weaker assumptions, by relying only on observations and removing the dependency on actions. The contributions in this work are complementary (and compatible) to \cite{torabi2018generative}.





\section{Background}


\subsection{Markov Decision Process}

We consider problems that satisfy the definition of a Markov Decision Process (MDP), formalized by the tuple: $(\mathcal{S}, \mathcal{A}, p(s), p(s'|s,a), r(s,a,s'), \gamma)$. Here $\mathcal{S}$, $\mathcal{A}$ represent the state and action spaces respectively, $p(s)$ is the initial state distribution, $p(s'|s,a)$ defines environment dynamics represented as a conditional state distribution, $r(s,a,s')$ is reward function and $\gamma$ the return discount factor.

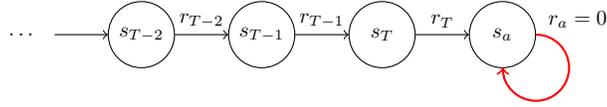
\begin{figure}
\label{fig:abs}
\centering
    \begin{tikzpicture}[transform shape, scale=0.8]
        \node[shape=circle,draw=white,minimum size=1.1cm] (X) at (-2,0) {$\ldots$};
        \node[shape=circle,draw=black,minimum size=1.1cm] (A) at (0,0) {$s_{T-2}$};
        \node[shape=circle,draw=black,minimum size=1.1cm] (B) at (2,0) {$s_{T-1}$};
        \node[shape=circle,draw=black,minimum size=1.1cm] (C) at (4,0) {$s_T$};
        \node[shape=circle,draw=black,minimum size=1.1cm] (D) at (6,0) {$s_a$};
        \draw [->](X) edge node { } (A);
        \draw [->](A) edge node[above] {$r_{T-2}$} (B);
        \draw [->](B) edge node[above] {$r_{T-1}$} (C);
        \draw [->](C) edge node[above] {$r_T$} (D);
        \draw[red,thick,->] (D.0) arc (90:90-270:5.5mm);
        \node at (7.25,0.25) {$r_a=0$};
    \end{tikzpicture}
\caption{Absorbing states for episodic tasks.}
\end{figure}

In continuing tasks, where environment interactions are unbounded in sequence length,
the returns for a trajectory $\tau=\{(s_t, a_t)\}_{t=0}^{\infty}$, are defined as $R_t = \sum_{k=t}^{\infty} \gamma^{k-t} r(s_k, a_k, s_{k+1})$.
In order to use the same notation for episodic tasks, whose finite length episodes end when reaching a terminal state, we can define a set of \emph{absorbing states} $s_a$ \citep{sutton1998reinforcement} that an agent enters after the end of episode, has zero reward and transitions to itself for all agent actions: $s_a \sim p(\cdot|s_T,a_T)$, $r(s_a, \cdot, \cdot)=0$ and $s_a \sim
 p(\cdot|s_a,\cdot)$ (see Figure \ref{fig:abs}). 
 With this above absorbing state notation, returns can be defined simply as $R_t = \sum_{k=t}^{T} \gamma^{k-t} r(s_k, a_k, s_{k+1})$. In reinforcement learning, the goal is to learn a policy that maximizes expected returns. 
 
In many imitation learning and IRL algorithms a common assumption is to assign zero reward value, often implicitly, to absorbing states.
As we will discuss in detail in Section \ref{sec:unbias}, our DAC algorithm will assign a learned, potentially non-zero reward for absorbing states and we will demonstrate empirically in Section \ref{sec:bugs}, that it is extremely important to properly handle the absorbing states for algorithms where rewards are learned.


\subsection{Adversarial Imitation Learning}


In order to learn a robust reward function we use the GAIL framework \citep{ho2016generative}. Inspired by maximum entropy IRL~\citep{ziebart2008maximum} and Generative Adversarial Networks (GANs)~\citep{goodfellow2014generative}, GAIL trains a binary classifier, $D(s,a)$, referred to as the \emph{discriminator}, to distinguish between transitions sampled from an expert and those generated by the trained policy. In standard GAN frameworks, a generator gradient is calculated by backprop through the learned discriminator. However, in GAIL the policy is instead provided a \emph{reward} for confusing the discriminator, which is then maximized via some on-policy RL optimization scheme (e.g. TRPO~\citep{schulman2015trust}):

\begin{equation}
    \max_\pi \max_D    \mathbb{E}_\pi[\log(D(s, a))] + \mathbb{E}_{\pi_E} [\log(1 - D(s, a))] - \lambda H(\pi) 
\label{eqn:gail}
\end{equation}
where $H(\pi)$ is an entropy regularization term.

The rewards learned by GAIL might not correspond to a true reward \citep{fu2017learning} but can be used to match the expert occupancy measure, which is defined as $\rho_{\pi_E}(s, a)=\sum_{t=0}^\infty \gamma^t p(s_t=s, a_t=a | \pi_E) $. \cite{ho2016generative} draw analogies between distribution matching using GANs and occupancy matching with GAIL. They demonstrate that by maximizing the above reward, the algorithm matches occupancy measures of the expert and trained policies with some regulation term defined by the choice of GAN loss function. 

In principle, GAIL can be incorporated with any on-policy RL algorithm. However, in this work we adapt it for off-policy training (discussed in Section~\ref{sec:offpolicygail}). As can be seen from Equation \ref{eqn:gail}, the algorithm requires state-action pairs to be sampled from the learned policy. In Section \ref{sec:algo} we will discuss what modifications are necessary to adapt the algorithm to off-policy training. 

\section{Discriminator-Actor-Critic}

In this section we will present the Discriminator-Actor-Critic (DAC) algorithm. This algorithm is comprised of two parts: a method for unbiasing adversarial reward functions, discussed in Section~\ref{sec:unbias}, and an off-policy discriminator formulation of AIL, discussed in Section~\ref{sec:offpolicygail}. A high level pictorial representation of this algorithm is also shown in Figure~\ref{fig:opgail}. The algorithm is formally summarized in Appendix~\ref{sec:appendix_algo}.

\subsection{Bias in Reward Functions}

In the following section, we present examples of bias present in reward functions in different AIL algorithms:
\begin{itemize}
    \item In the GAIL framework, and follow-up methods, such as GMMIL~\citep{kim2018imitation} and AIRL, zero reward is implicitly assigned for absorbing states, while some reward function, $r(s,a)$, assigns rewards for intermediate states depending on properties of a task.

    \item For certain environments, a survival bonus in the form of per-step positive reward is added to the rewards received by the agent. This encourages agents to survive longer in the environment to collect more rewards. We observe that a commonly used form of the reward function: $r(s,a) = -\log(1 - D(s,a))$ has worked well for environments that require a survival bonus. However, since the recovered reward function can never be negative, it cannot recover the true reward function for environments where an agent is required to solve the task as quickly as possible. Using this form of the reward function will lead to sub-optimal solutions. The agent is now incentivized to move in loops or take small actions (in continuous action spaces) that keep it close to the states in the expert's trajectories. The agent keeps collecting positive rewards without actually attempting to solve the task demonstrated by the expert (see Section \ref{sec:rewardbias}).\footnote{Note that this behavior was described in the early reward shaping literature \citep{ng1999policy}.}
    
    \item Another reward formulation is $r(s,a) = \log(D(s,a))$. This is often used for tasks with a per step penalty, when a part of a reward function consists of a negative constant assigned unconditionally of states and actions. However, this variant assigns only negative rewards and cannot learn a survival bonus. 
    Such strong priors might lead to good results even with no expert trajectories (as shown in Figure~\ref{fig:bias}).
    
\end{itemize}

From an end-user's perspective, it is undesirable to have to craft a different reward function for every new task. In the next section, we describe an illustrative example of a typical failure of biased reward functions. We also propose a method to unbias the reward function in our imitation learning algorithm such that it is able to recover different reward functions without adjusting the form of reward function.

\begin{figure}
\centering
    \begin{tikzpicture}[scale=0.85, transform shape]
        \node[shape=circle,draw=black] (A) at (0,0) {$s_1$};
        \node[shape=circle,draw=black] (B) at (2,0) {$s_2$};
        \node[shape=circle,draw=black] (C) at (4,0) {$s_g$};
        \draw[bend left,->] (A) edge node[above] {$a_{1 \rightarrow 2}$} (B);
        \draw[bend left,->] (B) edge node[below] {$a_{2 \rightarrow 1}$} (A);
        \path[->] (B) edge node[above] {$a_{2 \rightarrow g}$} (C);
        \node[below=1cm] at (2, 0) {a)};
        \node[shape=circle,draw=black] (D) at (5,0) {$s_1$};
        \node[shape=circle,draw=black] (E) at (7,0) {$s_2$};
        \node[shape=circle,draw=black] (F) at (9,0) {$s_g$};
        \path[->] (D) edge node[above=0.2cm] {$1: a_{1 \rightarrow 2}$} (E);
        \path[->] (E) edge node[above=0.2cm] {$2: a_{2 \rightarrow g}$} (F);
        \node[below=1cm] at (7, 0) {b)};
        \node[shape=circle,draw=black] (AA) at (10,0) {$s_1$};
        \node[shape=circle,draw=black] (AB) at (12,0) {$s_2$};
        \node[shape=circle,draw=black] (AC) at (14,0) {$s_g$};
        \draw[bend left,->] (AA) edge node[above] {$1,3: a_{1 \rightarrow 2}$} (AB);
        \draw[bend left,->] (AB) edge node[below] {$2: a_{2 \rightarrow 1}$} (AA);
        \path[->] (AB) edge node[above=0.2cm] {$4: a_{2 \rightarrow g}$} (AC);
        \node[below=1cm] at (12, 0) {c)};
    \end{tikzpicture}
    \label{fig:mdp}
\caption{a) An MDP with 3 possible states and 3 possible actions. b) Expert demonstration. c) A policy (potentially) more optimal than the expert policy according to the GAIL reward function.}
\end{figure}
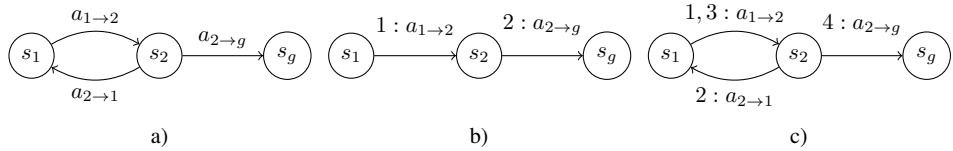

\label{sec:rewardbias}
\subsubsection{An Illustrative Example of Reward Bias}

Firstly, we illustrate how  $r(s,a) = -\log(1 - D(s,a))$ 
cannot match the expert trajectories with environments with per-step penalties.
Consider a simple MDP with 3 states: $s_1$, $s_2$, $s_{g}$, where $s_{g}$ is a goal state and agents receive a reward by reaching the goal state, and 3 actions: $a_{1 \rightarrow 2}, a_{2 \rightarrow 1}, a_{2 \rightarrow g}$; where $a_{i \rightarrow j}$ is such that $s_j \sim p(\cdot | s_i, a_{i \rightarrow j})$, as shown in Figure \ref{fig:mdp} a).
And for every state the expert demonstration is the following: $\pi_{E}(s_1)=a_{1 \rightarrow 2}$, $\pi_{E}(s_2)=a_{2 \rightarrow g}$, (as shown in Figure \ref{fig:mdp} b), and which clearly reaches the goal state in the optimal number of steps. Now consider the trajectory of Figure \ref{fig:mdp} c): $(s_1, a_{1 \rightarrow 2})  \rightarrow (s_2, a_{2 \rightarrow 1})
\rightarrow (s_1, a_{1 \rightarrow 2})
\rightarrow (s_2, a_{2 \rightarrow g})$. 
This trajectory will have the return $R_\pi = r(s_1, a_{1 \rightarrow 2}) + \gamma r(s_2, a_{2 \rightarrow 1}) + \gamma^2 r(s_1, a_{1 \rightarrow 2}) + \gamma^3 r(s_2, a_{2 \rightarrow g})$. While the expert return is $R_E = r(s_1, a_{1 \rightarrow 2}) + \gamma r(s_2, a_{2 \rightarrow g})$.

Assuming that we have a discriminator trained to convergence, it will assign $r(s_2, a_{2 \rightarrow 1})$ a value that is close to zero, since it never appears in expert demonstrations. Therefore, from 
$R_\pi < R_E$
one can derive
$r(s_1, a_{1 \rightarrow 2})  < \frac{(1 - \gamma^2)}{\gamma} r(s_2, a_{2 \rightarrow g})$.
Thus, for the loopy trajectory to have a smaller return than our expert policy, we need $r(s_1, a_{1 \rightarrow 2}) < \frac{0.0199}{0.99} \cdot r(s_2, a_{2 \rightarrow g})$, if $\gamma=0.99$ (a standard value). However, the optimal values for GAN discriminator in this case are $r(s_1, a_{1 \rightarrow 2}) = -log(1-0.5) \approx 0.3$ and 
$r(s_2, a_{2 \rightarrow g})=-log(1-2/3) \approx 0.477$. Hence, the inequality above does not hold. As such, the convergence of GAIL to the expert policy with this reward function is possible under only certain values of $\gamma$, and this value depends heavily on the task MDP. At the same time, since the reward function is strictly positive it implicitly provides a survival bonus. In other words, regardless of how the discriminator actually classifies state-action tuples, it always rewards the policy for avoiding absorbing states (see Section~\ref{sec:rew_func}).
Fundamentally, this characteristic makes it difficult to attribute policy performance to the robustness of the GAIL learned reward since the RL optimizer can often solve the task as long as the reward is strictly positive.

Another common reward variant, $r(s,a)=\log(D(s,a))$, which corresponds to the original saturating loss for GANs, penalizes every step and leads to collapsing in environments with a survival bonus. This phenomenon can be demonstrated using a reasoning similar to the one stated above.


Finally, AIRL uses the reward function: $r(s,a,s') = \log(D(s,a,s') -\log(1 - D(s,a,s'))$, which can assign both positive and negative rewards for each time step. In AIRL, 
as in the original GAIL, the agent receives zero reward at the end of the episode, leading to sub-optimal policies (and training instability) in environments with a survival bonus. In the beginning of training this reward function assigns rewards with a negative bias because it is easy for the discriminator to distinguish samples for an untrained policy and an expert, and so it is common for learned agents to finish an episode earlier (to avoid additional negative penalty) instead of trying to imitate the expert.


\label{sec:bugs}

\subsection{Unbiasing Reward Functions}

\label{sec:unbias}

In order to resolve the issues described in Section \ref{sec:bugs}, we suggest explicitly learning rewards for absorbing states for expert demonstrations and trajectories produced by a policy.
Thus, the returns for final states are defined now $R_T = r(s_T, a_T) + \sum_{t=T+1}^\infty \gamma^{t-T} r(s_a, \cdot)$ with a learned reward $r(s_a, \cdot)$ instead of just $R_T = r(s_T, a_T)$.

We implemented these absorbing states by adding an extra indicator dimension that indicates whether the state is absorbing or not, for absorbing states we set the indicator dimension to one and all other dimensions to zero. The GAIL discriminator can distinguish whether reaching an absorbing state is a desirable behavior from the expert's perspective and assign the rewards accordingly.

Instead of recursively computing the Q values, this issue can be addressed by analytically deriving the following returns for the terminal states: $R_T = r(s_T, a_T) + \frac{\gamma r(s_a, \cdot)}{1-\gamma}$.
However, in practice this alternative was much less stable.


\subsection{Addressing Sample Inefficiency}




\label{sec:offpolicygail}

As previously mentioned, GAIL requires a significant number of interactions with a learning environment in order to imitate an expert policy. To address the sample inefficiency of GAIL, we use an off-policy RL algorithm
and perform off-policy training of the GAIL discriminator performed in the following way: instead of sampling trajectories from a policy directly, we sample transitions from a replay buffer $\mathcal{R}$ collected while performing off-policy training:

\begin{equation}
    \max_D    \mathbb{E}_{\mathcal{R}}[\log(D(s, a))] + \mathbb{E}_{\pi_E} [\log(1 - D(s, a))] - \lambda H(\pi).
    \label{eqn:offpolicyd}
\end{equation}

Equation \ref{eqn:offpolicyd} tries to match the occupancy measures between the expert and the distribution induced by the replay buffer $\mathcal{R}$, which can be seen as a mixture of all policy distributions that appeared during training, instead of the latest trained policy $\pi$. In order to recover the original on-policy expectation, one needs to use importance sampling:

\begin{equation}
    \max_D    \mathbb{E}_{\mathcal{R}}\left[\frac{p_{\pi_\theta}(s,a)}{p_{\mathcal{R}}(s,a)}\log(D(s, a))\right] + \mathbb{E}_{\pi_E} [\log(1 - D(s, a))] - \lambda H(\pi).
\end{equation}

However, it can be challenging to properly estimate these densities and the discriminator updates might have large variance.
We found that the algorithm works well in practice with the importance weight omitted.

We use the GAIL discriminator in order to define rewards for training a policy using TD3; we update per-step rewards every time when we pull transitions from the replay buffer using the latest discriminator.
The TD3 algorithm provides a good balance between sample complexity and simplicity of implementation and so is a good candidate for practical applications. Additionally, depending on the distribution of expert demonstrations and properties of the task, off-policy RL algorithms can effectively handle multi-modal action distributions; for example, this can be achieved for the Soft Actor Critic algorithm~\citep{haarnoja2018soft} using the reparametrization trick~\citep{kingma2014adam} with a normalizing flow \citep{rezende2015variational} as described in \cite{haarnoja2018latent}.


\label{sec:algo}


\section{Experiments}
\tompson{TODO for ICLR camera ready. Plots need work: random is white. "m" --> "M", etc. LOTS of cleanup.}

\label{sec:exp}

We implemented the DAC algorithm described in Section~\ref{sec:algo} using TensorFlow Eager~\citep{tensorflow2015-whitepaper} and we evaluated it on popular benchmarks for continuous control simulated in MuJoCo~\citep{todorov2012mujoco}. We also define a new set of robotic continuous control tasks (described in detail below) simulated in PyBullet~\citep{coumans2016pybullet}, and a Virtual Reality (VR) system for capturing human examples in this environment; human examples constitute a particularly challenging demonstration source due to their noisy, multi-modal and potentially sub-optimal nature, and we define episodic multi-task environments as a challenging setup for adversarial imitation learning.

For the critic and policy networks we used the same architecture as in \cite{fujimoto2018addressing}: a 2 layer MLP with ReLU activations and 400 and 300 hidden units correspondingly. We also add gradient clipping~\citep{pascanu2013difficulty} to the actor network with clipping value of 40. For the discriminator we used the same architecture as in \cite{ho2016generative}: a 2 layer MLP with 100 hidden units and $\tanh$ activations. We trained all networks with the Adam optimizer~\citep{kingma2014adam} and decay learning rate by starting with initial learning rate of $10^{-3}$ and decaying it by 0.5 every $10^5$ training steps for the actor network.

\tompson{I don't think Sergey's ``research questions" comment was actually addressed.}

In order to make the algorithm more stable, especially in the off-policy regime when the discriminator can easily over-fit to training data, we use regularization in the form of gradient penalties~\citep{gulrajani2017improved} for the discriminator.
Originally, this was introduced as an alternative to weight clipping for Wasserstein GANs~\citep{arjovsky2017wasserstein}, but later it was shown that it helps to make JS-based GANs more stable as well~\citep{lucic2017gans}.



We replicate the experimental setup of \cite{ho2016generative}: expert trajectories are sub-sampled by retaining every 20 time steps starting with a random offset (and fixed stride). It is worth mentioning that, as in \cite{ho2016generative}, this procedure is done in order to make the imitation learning task harder. With full trajectories, behavioral cloning provides competitive results to GAIL.

\tompson{TODO for camera ready: Add experiments in the appendix for this.}

Following \cite{henderson2017deep} and \cite{fujimoto2018addressing}, we perform evaluation using 10 different random seeds. For each seed, we compute average episode reward using 10 episodes and running the policy without random noise. As in \cite{ho2016generative} we plot reward normalized in such a way that zero corresponds to a random reward while one corresponds to expert rewards. We compute mean over all seeds and visualize half standard deviations. In order to produce the same evaluation for GAIL we used the original implementation\footnote{\href{https://github.com/openai/imitation}{https://github.com/openai/imitation}} of the algorithm.

\subsection{Off Policy DAC Algorithm}

\begin{figure}
\centering

\includegraphics[width=0.31\textwidth]{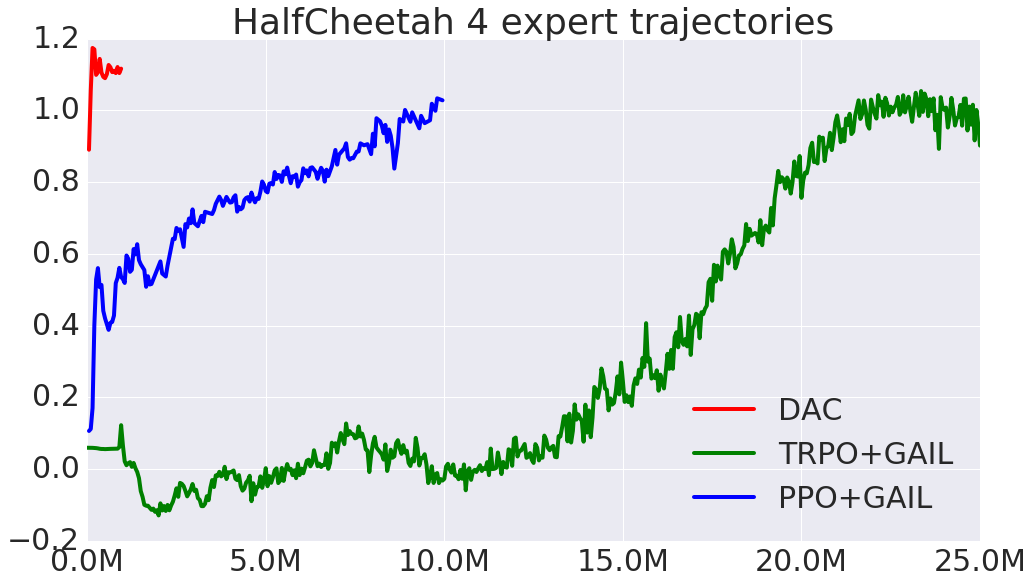}
\includegraphics[width=0.31\textwidth]{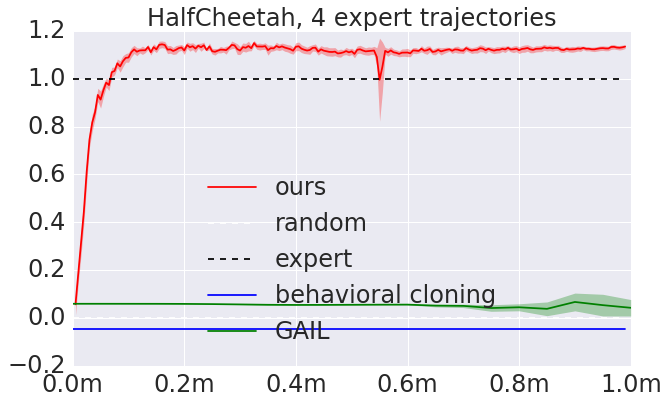}
\includegraphics[width=0.31\textwidth]{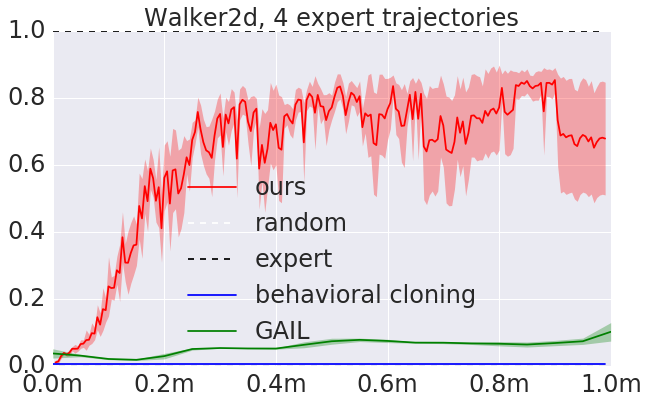}
\\
\includegraphics[width=0.31\textwidth]{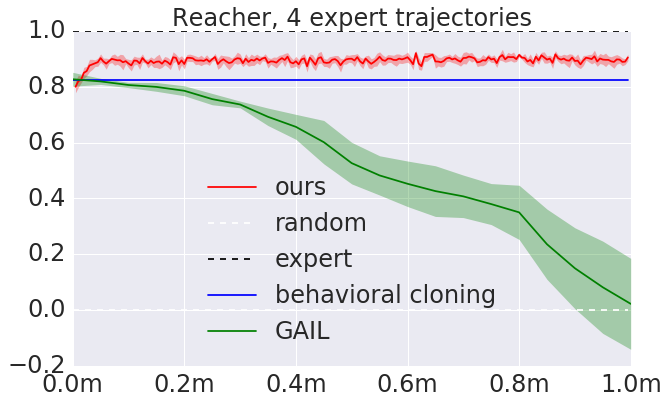}
\includegraphics[width=0.31\textwidth]{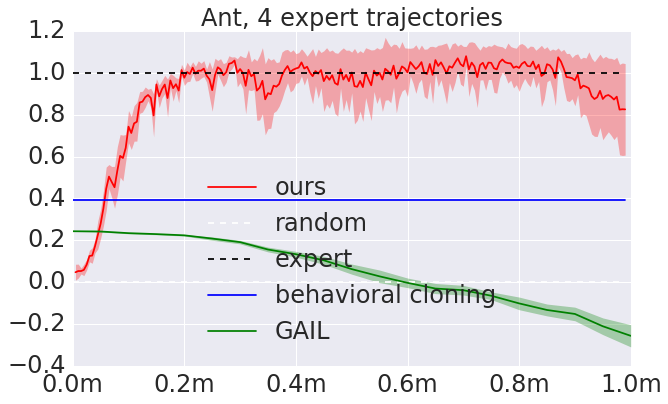}
\includegraphics[width=0.31\textwidth]{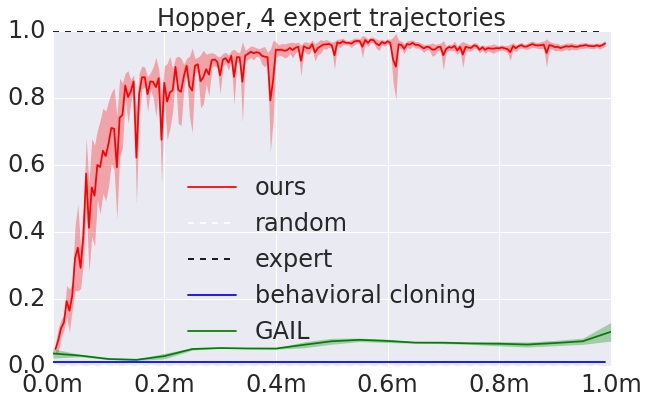}

\caption{Comparisons of algorithms using 4 expert demonstrations. y-axis corresponds to normalized reward (0 corresponds to a random policy, while 1 corresponds to an expert policy).}
\label{fig:results}
\end{figure}

Evaluation results of the DAC algorithm on a suite of MuJoCo tasks are shown in Figure~\ref{fig:results}, as are the GAIL (TRPO) and BC basline results. In the top-left plot, we show DAC is an order of magnitude more sample efficent than then TRPO and PPO based GAIL baselines. In the other plots, we show that by using a significantly smaller number of environment steps (orders of magnitude fewer), our DAC algorithm reaches comparable expected reward as the GAIL baseline. Furthermore, DAC outperforms the GAIL baseline on all environments within a 1 million step threshold.
A comprehensive suit of results can be found in Appendix \ref{sec:appendix_results}, Figure \ref{fig:appendix_results}.

\subsection{Reward Bias}
\label{sec:rew_func}

\begin{figure}
\centering
\includegraphics[width=0.31\textwidth]{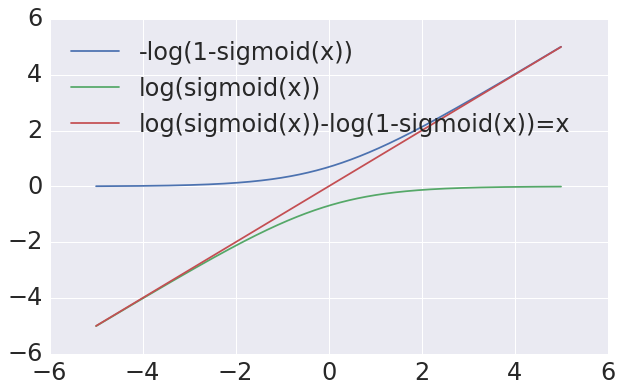}
\includegraphics[width=0.31\textwidth]{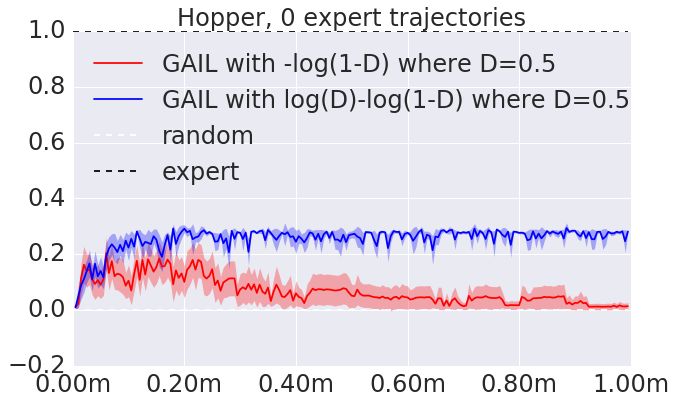}
\caption{Reward functions that can be used in GAIL (left).
Even without training some reward functions can perform well on some tasks (right).}
\label{fig:bias}
\end{figure}

As discussed in Section~\ref{sec:bugs}, the reward function variants used with GAIL can have implicit biases when used without handling absorbing states. Figure~\ref{fig:bias} demonstrates how bias affects results on an environment with survival bonus when using the reward function of \cite{ho2016generative}: $r(s,a)=-\log(1 - D(s,a))$. Surprisingly, when using a fixed and untrained GAIL discriminator that outputs 0.5 for every state-action pair, we were able to reach episode rewards of around 1000 on the Hopper environment, corresponding to approximately one third of the expert performance. Without any reward learning, and using no expert demonstrations, the agent can learn a policy that outperforms behavioral cloning (Figure \ref{fig:bias}). Therefore, the choice of a specific reward function might already provide strong prior knowledge that helps the RL algorithm to move towards recovering the expert policy, irrespective of the quality of the learned reward. 

\kostrikov{we also need to say that it's easy to design a problem that has survival bonus for some states and step penalty for the others. So it's even impossible to design the right reward.}

\begin{figure}
\centering
\includegraphics[width=0.31\textwidth]{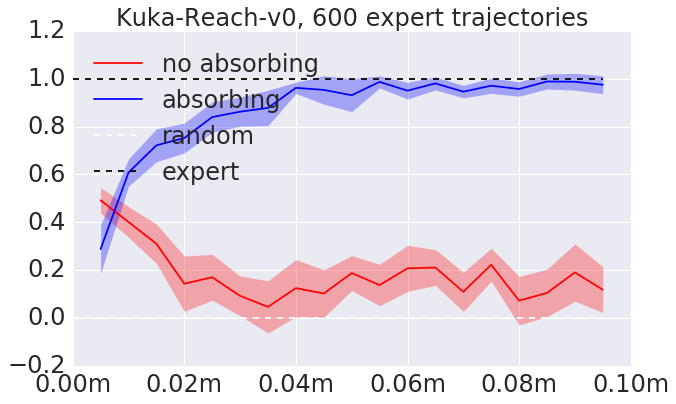}
\includegraphics[width=0.31\textwidth]{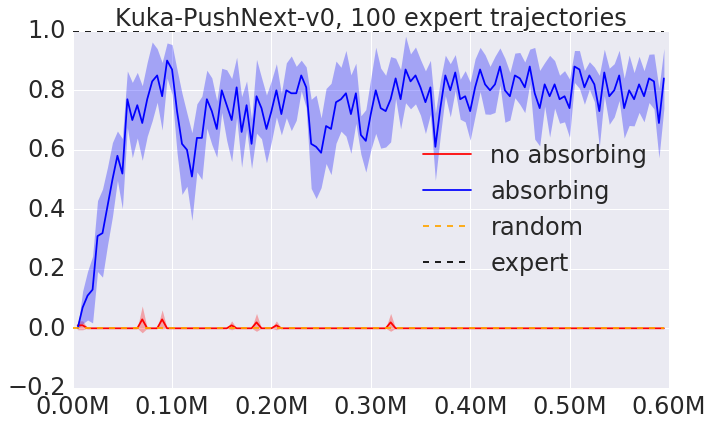}
\caption{Effect of absorbing state handling on Kuka environments with human demonstrations.}
\label{fig:bias3}
\end{figure}

Additionally, we evaluated our method on two environments with per-step penalty (see Figure~\ref{fig:bias3}). These environment are simulated in PyBullet and consist of a Kuka IIWA arm and 3 blocks on a virtual table. A rendering of the environment can be found in Appendix~\ref{sec:kuka_pics}, Figure~\ref{fig:kuka_pics}. Using a Cartesian displacement action for the gripper end-effector and a compact observation-space (consisting of each block's 6DOF pose and the Kuka's end-effector pose), the agent must either a) reach one of the 3 blocks in the shortest number of frames possible (the target block is provided to the policy as a one-hot vector), which we call \emph{Kuka-Reach}, or b) push one block along the table so that it is adjacent to another block, which we call \emph{Kuka-PushNext}. For evaluation, we define a sparse reward indicating successful task completion (within some threshold). For these imitation learning experiments, we use human demonstrations collected with a VR setup, where the participant wears a VR headset and controls in real-time the gripper end-effector using a 6DOF controller.

Using the reward defined as $r(s,a) = -log(1-D(s,a))$ and without absorbing state handling, the agent completely fails to recover the expert policy given $600$ expert trajectories without sub-sampling (as shown in Figure~\ref{fig:bias}). In contrast, our DAC algorithm quickly learns to imitate the expert, despite using noisy and potentially sub-optimal human demonstrations.

\begin{figure}
\centering
\includegraphics[width=0.36\textwidth]{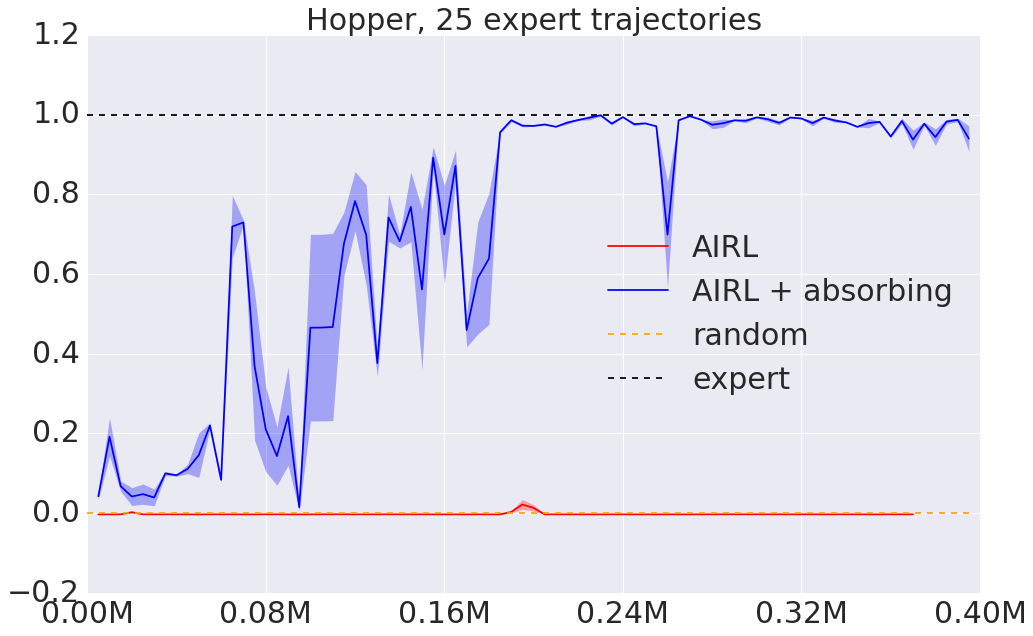}
\includegraphics[width=0.36\textwidth]{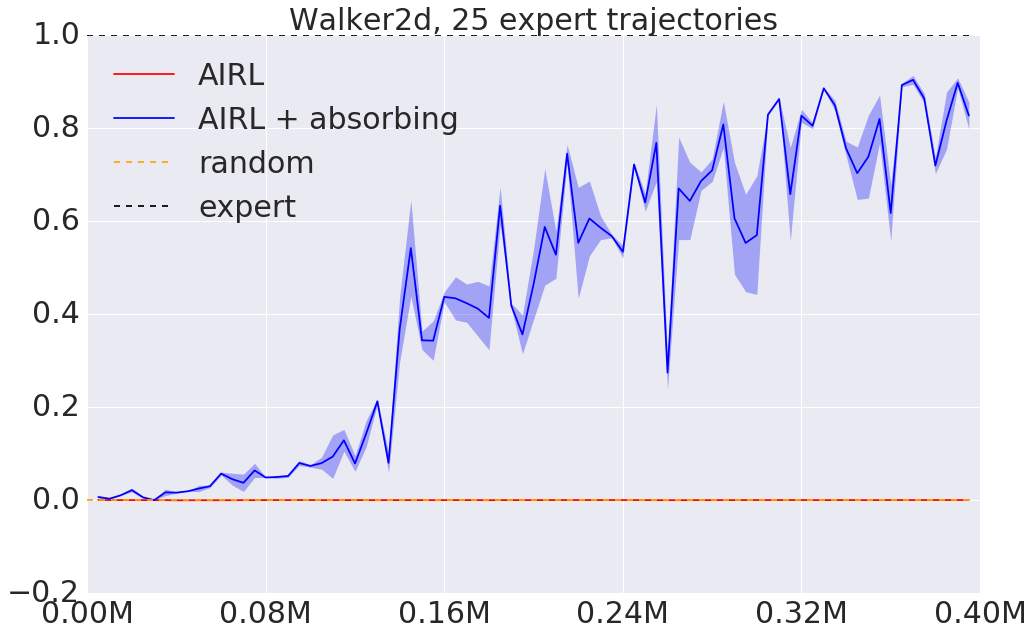}

\caption{Effect of learning absorbing state rewards when using an AIRL discriminator within the DAC Framework.}
\label{fig:bias2}
\end{figure}

As discussed, alternative reward functions do not have this positive bias but still require proper handling of the absorbing states as well in order to avoid early termination due to incorrectly assigned per-frame penalty. Figure \ref{fig:bias2} illustrates results for AIRL with and without learning rewards for absorbing states. For these experiments we use the discriminator structure from \cite{fu2017learning} in combination with the TD3 algorithm.

\section{Conclusion}

In this work we address several important issues associated with the popular GAIL framework. In particular, we address 1) sample inefficiency with respect to policy transitions in the environment and 2) we demonstrate a number of reward biases that can either implicitly impose prior knowledge about the true reward, or alternatively, prevent the policy from imitating the optimal expert. To address reward bias, we propose a simple mechanism whereby the rewards for absorbing states are also learned, which negates the need to hand-craft a discriminator reward function for the properties of the task at hand. In order to improve sample efficiency, we perform off-policy training of the discriminator and use an off-policy RL algorithm. We show that our algorithm reaches state-of-the-art  performance for an imitation learning algorithm on several standard RL benchmarks, and is able to recover the expert policy given a significantly smaller number of samples than in recent GAIL work. We will make the code for this project public following review.



\bibliography{iclr2019_conference}
\bibliographystyle{iclr2019_conference}


\newpage
\appendix

\section{DAC Algorithm}

\label{sec:appendix_algo}

{\centering
\begin{minipage}{\linewidth}
\begin{small}
\begin{algorithm}[H]
    \textbf{Input}: expert replay buffer $\mathcal{R}_E$
    \caption{Discriminative-Actor-Critic Adversarial Imitation Learning Algorithm}
    \begin{algorithmic}
    \Procedure{WrapForAbsorbingStates}{$\tau$}
        \If {$s_T$ is a terminal state}
            \State $(s_T, a_T,\cdot, s'_T) \gets (s_T, a_T, \cdot, s_a)$
            \State $\tau \gets \tau \cup \{(s_a, \cdot, \cdot, s_a)\}$
        \EndIf
    \State \Return $\tau$
    \EndProcedure
    \State \vspace{0.2cm}
    \State Initialize replay buffer $\mathcal{R} \gets \emptyset$
    \For { $\tau=\{(s_t,a_t,\cdot, s'_t)\}_{t=1}^T \in  \mathcal{R}_E $}
    \State $\tau \gets$ WrapForAbsorbingState$(\tau)$  \Comment{Wrap expert rollouts with absorbing states}
    \EndFor
    \For{$n=1,\ldots,$}
        \State Sample $\tau=\{(s_t, a_t, \cdot, s'_t)\}_{t=1}^T$ with $\pi_\theta$
        \State $\mathcal{R}  \gets \mathcal{R} \cup$ WrapForAbsorbingState$(\tau)$
        \Comment Update Policy Replay Buffer
        \For{$i=1,\ldots,|\tau|$}
            \State $\{(s_t,a_t,\cdot, \cdot)\}_{t=1}^B \sim \mathcal{R}$, \hspace{0.1cm} $\{(s'_t,a'_t,\cdot, \cdot)\}_{t=1}^B \sim \mathcal{R}_E$ 
            \Comment Mini-batch sampling
            \State $ \mathcal{L} = \sum_{b=1}^B \log D(s_b,a_b)-\log(1-D(s'_b,a'_b))$
            \State Update $D$ with GAN+GP
        \EndFor

        \For{$i=1,\ldots,|\tau|$}
            \State $\{(s_t,a_t,\cdot, s'_t)\}_{t=1}^B \sim \mathcal{R}$ 
            \For{$b=1,\ldots, B$}
                \State $r \gets  \log D(s_b,a_b)-\log(1-D(s_b,a_b))$
                \State $(s_b,a_b,\cdot, s'_b) \gets (s_b,a_b,r, s'_b)$
                \Comment Use current reward estimate.
            \EndFor
            \State Update $\pi_\theta$ with TD3
        \EndFor
    \EndFor
    \end{algorithmic}
    \label{algo:ours}
\end{algorithm}
\end{small}
\end{minipage}
\par
}

\newpage
\section{Supplementary Results on Mujoco Environments}
\label{sec:appendix_results}

\begin{figure}[H]
\centering

\includegraphics[width=0.31\textwidth]{images/results/HalfCheetah4.png}
\includegraphics[width=0.31\textwidth]{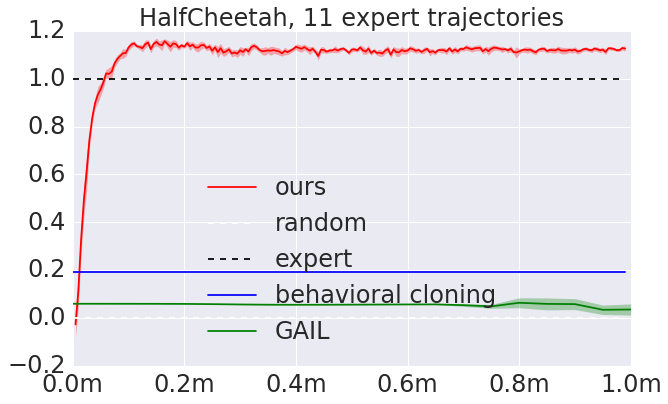}
\includegraphics[width=0.31\textwidth]{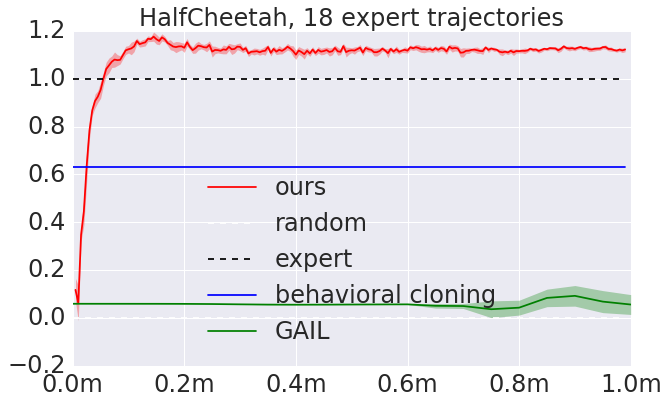}
\\
\includegraphics[width=0.31\textwidth]{images/results/Hopper4.png}
\includegraphics[width=0.31\textwidth]{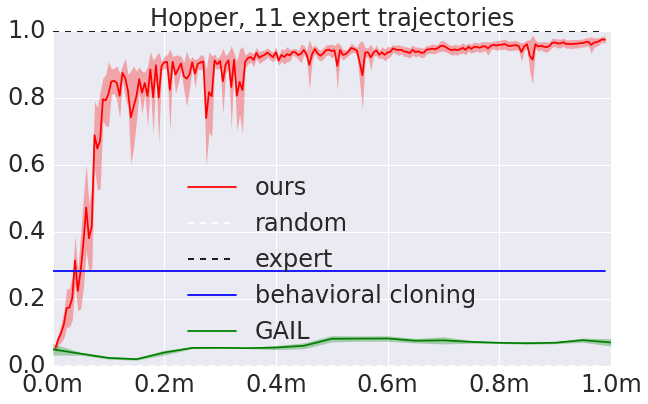}
\includegraphics[width=0.31\textwidth]{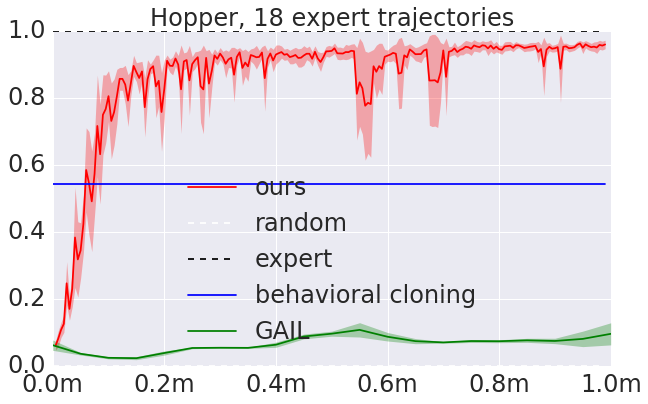}
\\
\includegraphics[width=0.31\textwidth]{images/results/Walker2d4.png}
\includegraphics[width=0.31\textwidth]{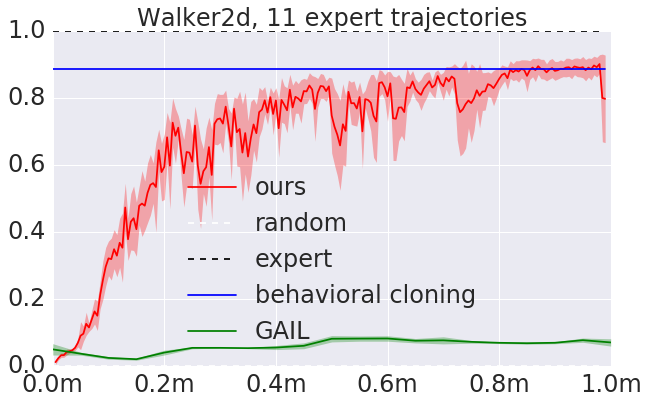}
\includegraphics[width=0.31\textwidth]{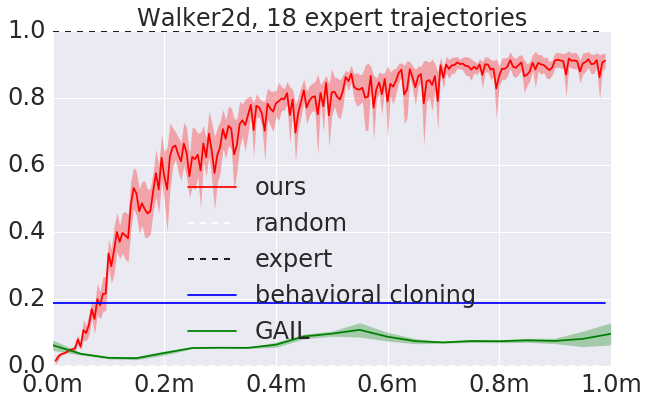}
\\
\includegraphics[width=0.31\textwidth]{images/results/Reacher4.png}
\includegraphics[width=0.31\textwidth]{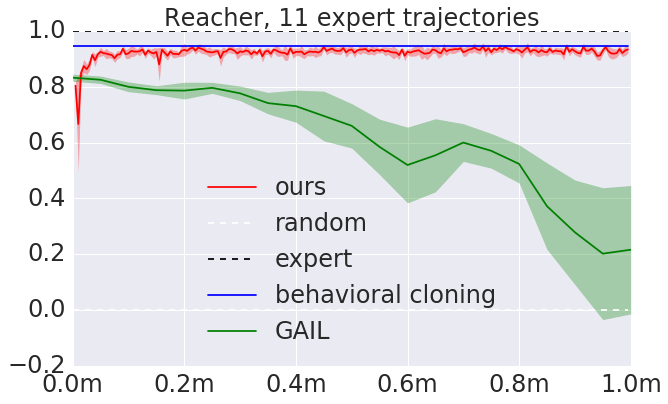}
\includegraphics[width=0.31\textwidth]{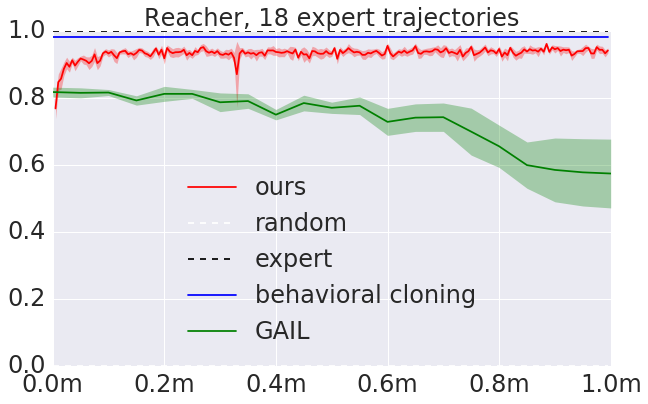}
\\
\includegraphics[width=0.31\textwidth]{images/results/Ant4.png}
\includegraphics[width=0.31\textwidth]{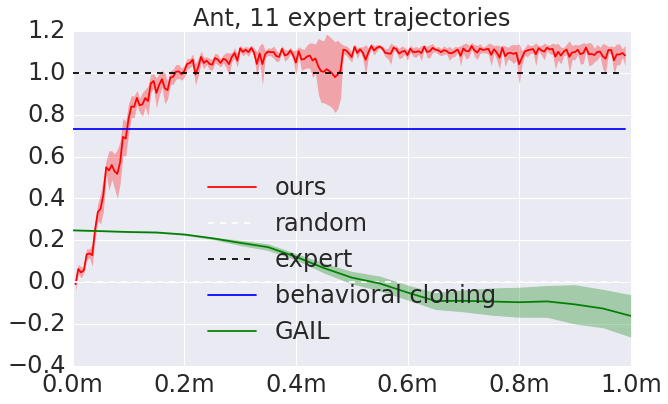}
\includegraphics[width=0.31\textwidth]{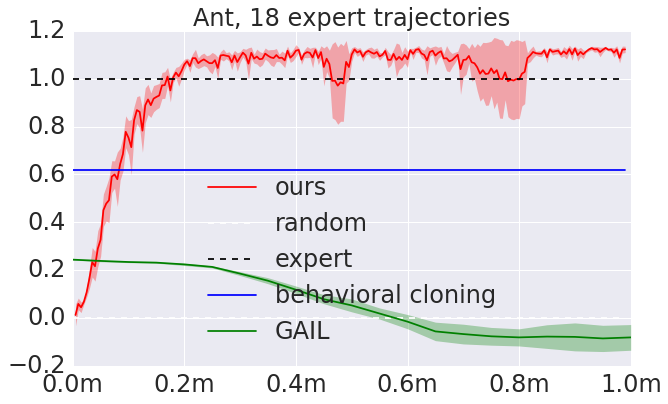} 
\\

\caption{Comparisons of different algorithms given the same number of expert demonstrations. y-axis corresponds to normalized reward (0 corresponds to a random policy, while 1 corresponds to an expert policy). \tompson{Add Kuka environments as 2 additional rows for camera ready.}}
\label{fig:appendix_results}
\end{figure}

\newpage
\section{Kuka-IIWA Simulated Environment}
\label{sec:kuka_pics}

\begin{figure}[H]
\centering

\includegraphics[width=0.4\textwidth]{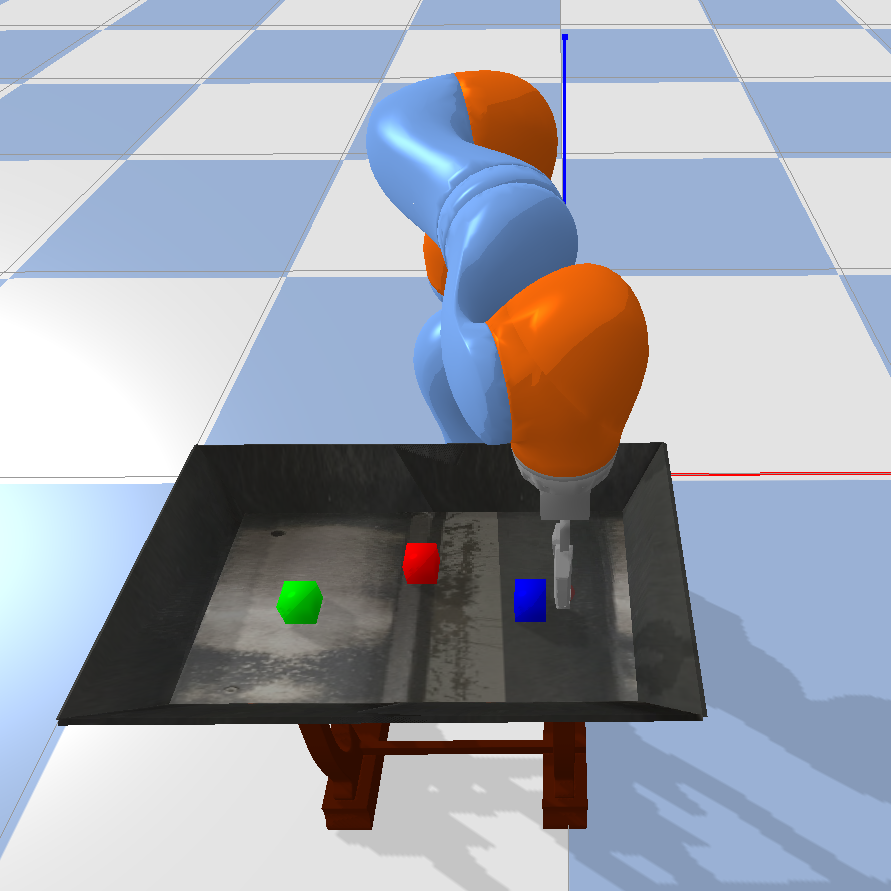} \hspace{0.1cm}
\includegraphics[width=0.4\textwidth]{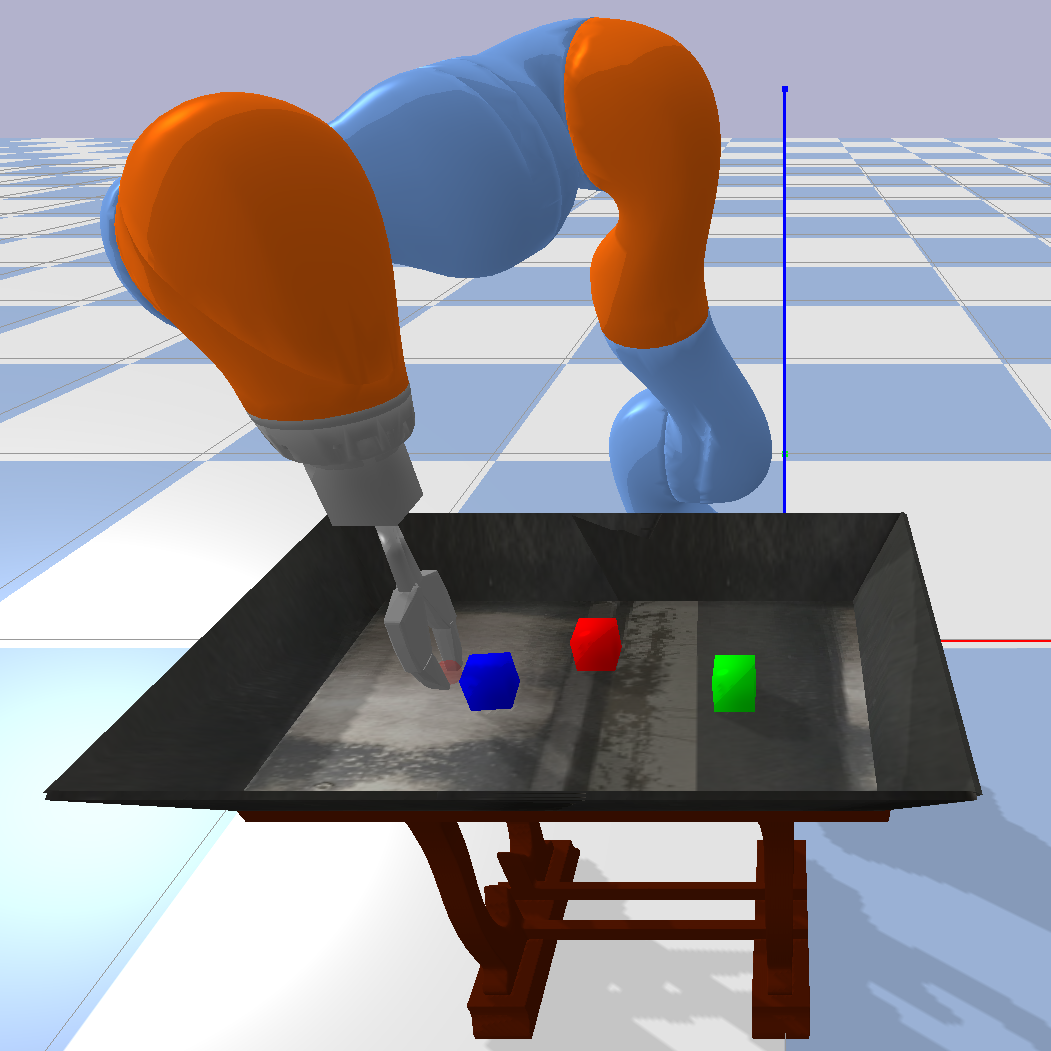}

\caption{Renderings of our Kuka-IIWA environment. Using a VR headset and 6DOF controller, a human participant can control the 6DOF end-effector pose in order to record expert demonstrations. In the \emph{Kuka-Reach} tasks, the agent must bring the robot gripper to 1 of the 3 blocks (where the state contains a 1-hot encoding of the task) and for the \emph{Kuka-PushNext} tasks, the agent must use the robot gripper to push one block next to another.}
\label{fig:kuka_pics}
\end{figure}

\end{document}